\documentclass[letterpaper]{article} 
\usepackage[preprint]{aaai2027}
\usepackage[hyphens]{url}  
\usepackage{graphicx} 
\usepackage{natbib}  
\usepackage{caption} 
\usepackage{booktabs}
\usepackage{amsmath}
\usepackage{amssymb}
\usepackage{tikz}
\usetikzlibrary{positioning,calc}
\usepackage{pgfplots}
\pgfplotsset{compat=1.17}
\newcommand{\system}{\textsc{StateAuditor}}
\newcommand{\stale}{\textsc{Stale}}
\newcommand{\horizonbench}{\textsc{HorizonBench}}
\newcommand{\valid}{\texttt{VALID}}
\newcommand{\staleflag}{\texttt{STALE}}
\newcommand{\unknown}{\texttt{UNKNOWN}}

\newcommand{\OffMacro}{.879}
\newcommand{\OffCILow}{.853}
\newcommand{\OffCIHigh}{.904}
\newcommand{\OffDraftMacro}{.375}

\newcommand{\OffFix}{616}
\newcommand{\OffBreak}{11}
\newcommand{\OffScenImproved}{331}
\newcommand{\OffScenRegressed}{1}
\newcommand{\OffPExp}{97}

\newcommand{\OffUsdPerScenario}{0.49}
\newcommand{\OffFailedCells}{145}

\newcommand{\OffLocalJudgeN}{400}

\newcommand{\OffBackboneDraft}{.390}
\newcommand{\OffBackboneFinal}{.573}

\newcommand{\OffBackbonePExp}{8}

\newcommand{\OffPriorMacro}{.680}

\newcommand{\OffLocalPriorVta}{.858}
\newcommand{\OffLocalPriorBase}{.717}

\newcommand{\OffLocalPriorWins}{125}

\newcommand{\OffLocalPriorLosses}{32}

\newcommand{\OffLocalPriorPExp}{13}

\newcommand{\HumanImproveA}{18}
\newcommand{\HumanImproveB}{21}
\newcommand{\HumanRegressA}{5}
\newcommand{\HumanRegressB}{1}

\newcommand{\HumanKappa}{0.126}

\newcommand{\StrictN}{400}
\newcommand{\StrictMacro}{.736}
\newcommand{\StrictCILow}{.707}
\newcommand{\StrictCIHigh}{.767}
\newcommand{\StrictBaseMacro}{.686}
\newcommand{\StrictDelta}{5.0}
\newcommand{\StrictDeltaLow}{2.9}
\newcommand{\StrictDeltaHigh}{7.2}
\newcommand{\StrictScenWins}{84}

\newcommand{\StrictScenLosses}{37}
\newcommand{\StrictScenPExp}{4}
\newcommand{\StrictIpaWins}{53}
\newcommand{\StrictIpaLosses}{6}
\newcommand{\StrictPrWins}{31}
\newcommand{\StrictPrLosses}{10}
\newcommand{\StrictSrWins}{24}
\newcommand{\StrictSrLosses}{32}
\newcommand{\StrictTTwoSrWins}{12}
\newcommand{\StrictTTwoSrLosses}{21}
\newcommand{\StrictTTwoIpaWins}{33}
\newcommand{\StrictTTwoIpaLosses}{2}

\newcommand{\RepairOnlyMacro}{.723}
\newcommand{\RepairOnlyWins}{83}
\newcommand{\RepairOnlyLosses}{48}
\newcommand{\RepairOnlyP}{2.8\times10^{-3}}
\newcommand{\RepairOnlyTTwoSrWins}{8}
\newcommand{\RepairOnlyTTwoSrLosses}{32}
\newcommand{\VerifyNRewrites}{162}

\newcommand{\VerifyPreservedN}{109}
\newcommand{\VerifyWeakenedN}{32}
\newcommand{\VerifyChangedN}{21}
\newcommand{\VerifyRelevantRate}{99}
\newcommand{\VerifyUnnecessaryRate}{94}

\newcommand{\VerifyTotalToWrong}{6}
\newcommand{\VerifyTotalFixed}{18}

\newcommand{\DirectMacro}{.692}
\newcommand{\DirectDelta}{0.6}
\newcommand{\DirectWins}{53}
\newcommand{\DirectLosses}{45}
\newcommand{\TransDelta}{4.4}
\newcommand{\TransWins}{52}
\newcommand{\TransLosses}{7}
\newcommand{\TransPExp}{9}
\newcommand{\TransDeltaDisj}{4.3}
\newcommand{\TransWinsDisj}{51}
\newcommand{\TransLossesDisj}{12}
\newcommand{\VarNRuns}{3}
\newcommand{\VarMean}{.733}
\newcommand{\VarStd}{0.002}

\newcommand{\VarMinAdv}{4.5}
\newcommand{\HardN}{120}

\newcommand{\HardMOnePred}{.542}
\newcommand{\HardMOneFull}{.533}

\newcommand{\HardSemPred}{.633}
\newcommand{\HardSemFull}{.600}

\newcommand{\HbFullBase}{.308}
\newcommand{\HbFullVta}{.458}
\newcommand{\HbFullStaleBase}{.29}
\newcommand{\HbFullStaleVta}{.20}

\newcommand{\HbFullWins}{23}
\newcommand{\HbFullLosses}{5}
\newcommand{\HbFullP}{.001}

\newcommand{\HbFullClustP}{.007}
\newcommand{\HbCtrlNotrans}{.417}
\newcommand{\HbCtrlVtaNotransWins}{10}
\newcommand{\HbCtrlVtaNotransLosses}{5}
\newcommand{\HbCtrlVtaNotransP}{.30}
\newcommand{\HbCtrlNotransBaseWins}{21}
\newcommand{\HbCtrlNotransBaseLosses}{8}

\newcommand{\GateRejectStruct}{900}
\newcommand{\GateStructTotal}{900}
\newcommand{\GateNonsupVerify}{290}
\newcommand{\GateNonsupN}{290}

\newcommand{\SemGateTrapVta}{18}
\newcommand{\SemGateTrapSem}{6}

\title{When Memory Updates but Behavior Does Not: Repairing\\
Implicit Stale Dependencies in Personalized Agent Responses}
\author{Haofei Sun\textsuperscript{\rm 1}, Lin He\textsuperscript{\rm 2}}
\affiliations{\textsuperscript{\rm 1}The University of Texas at Arlington\\
\textsuperscript{\rm 2}The University of Tennessee, Knoxville\\
haofei.sun@uta.edu, lynnhe@utk.edu}

\begin{document}

\maketitle

\begin{abstract}
Memory-augmented agents can know that a user's stored state is outdated
and still plan around the old value. The \stale{} benchmark calls this
the implicit policy adaptation (IPA) gap. We identify one structural
contributor: draft-anchored verification checks what a response
\emph{says}, and in an open-ended response the stale dependency is
usually unsaid. \system{} therefore audits in the opposite direction,
from stored state to draft. An LLM proposes candidate old$\to$new
transitions from timestamped evidence; deterministic code pins each quotation
to a single entry, checks that the new evidence really is newer, and
lets only these verified transitions trigger repair. What is verified is
provenance and chronology --- not semantic supersession. On \stale{}'s
full protocol (400 scenarios, 50-session histories, one independent
response per query), strict single-query VTA scores \StrictMacro{}
against \StrictBaseMacro{} for our locked predecessor under the same
judge: a $+\StrictDelta$-point paired gain (95\% CI
$[+\StrictDeltaLow,+\StrictDeltaHigh]$) coming almost entirely from
IPA and premise resistance (PR). The benchmark's own judge, from a
third model family,
reproduces the gain ($.738$ vs.\ $.680$). On an independent cross-family
preference-evolution benchmark (\horizonbench{}), the full
draft-audit-repair pipeline over a gold-derived structured store raises
current-preference accuracy (user-clustered $p{<}.01$), though a matched
control shows most of this external gain is the draft-side audit itself;
a harder authored lifecycle set gives no gain,
bounding the claim while false invalidation stays controlled (Limitations).
On \stale{}, by contrast, a matched control (same evidence, adapter,
and call budget) scores only \DirectMacro{} ($+\DirectDelta$ over the
predecessor, n.s.), attributing the \stale{} gain to the transition
machinery rather than added context or calls. We make no claim about
general-purpose agent memory.
\end{abstract}

\section{Introduction}

Personal AI assistants accumulate long-term memory, and that memory goes
stale. Later observations rarely explicitly negate earlier beliefs:
``setting up utilities in my new Austin apartment'' implicitly
invalidates ``lives in Seattle.'' The \stale{} benchmark
\citep{stale2026} shows that this implicit-conflict regime is
particularly difficult for memory-augmented agents. Systems may resolve
the user's current state correctly yet fail to adapt their open-ended
behavior, a failure termed the implicit policy adaptation (IPA) gap. The
benchmark's write-side prototype achieves 91\% accuracy on state
resolution but only 32\% on IPA, and the updated evidence is
visible in 67.8\% of failed IPA cases. Making an update visible to the
generator is therefore insufficient: write- and read-side memory
controls do not guarantee that the final response will use the current
state.

Auditing generated text has been studied extensively: RARR
\citep{gao2023rarr}, CoVe \citep{dhuliawala2024cove}, and trained
editors \citep{mishra2024fava} decompose an output into claims and check
them against evidence. These methods primarily verify static world
knowledge, and their draft-anchored decomposition is structurally
incomplete for IPA: an open-ended response may depend on a stale belief
without stating it, so claim extraction does not expose the relevant
premise, and increasing auditor scale yields only limited gains in
open-ended recall (Fig.~\ref{fig:blindness}). The missing direction is
\emph{state-to-draft} verification: for each changed attribute, test
whether the response is consistent with the attribute's current value,
including dependencies the response leaves unstated.

So we audit the response that is actually produced, not just what the
generator was shown. \system{} operates after a memory-augmented
generator drafts a response: it (i) extracts user-state premises both from the draft and from the
question's presuppositions; (ii) assigns
lifecycle-aware \valid{}/\staleflag{}/\unknown{} verdicts against
temporally ordered memory, attaching the current value; (iii) proposes
state transitions and deterministically verifies their provenance and
chronology; and (iv) regenerates under a typed directive ---
\emph{repair}, \emph{correct-and-inform}, or \emph{verify} (answer,
then ask).

Contributions. \textbf{(1) Finding.} We identify \emph{draft-anchored
verification blindness}: in open-ended personalized requests, stale
assumptions often appear as implicit dependencies rather than explicit
claims. Across auditor scales, stale-premise recall decreases from
$0.44$--$1.0$ on explicit and presupposed probes to $0.06$--$0.38$ on
open-ended probes
(Fig.~\ref{fig:blindness}).
\definecolor{figmodelb}{HTML}{2563EB}
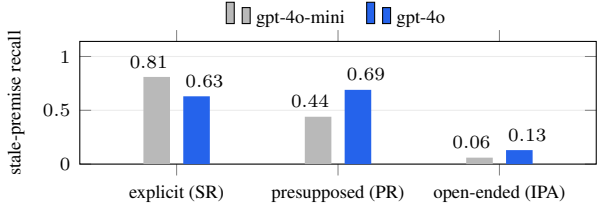
\begin{figure}[t]
\centering
\begin{tikzpicture}
\begin{axis}[
  width=\columnwidth, height=3.2cm,
  ybar=5pt, bar width=10pt, ymin=0, ymax=1.14,
  enlarge x limits=0.3,
  ylabel={stale-premise recall}, ylabel near ticks,
  y label style={font=\scriptsize},
  ytick={0,.5,1}, y tick label style={font=\scriptsize},
  symbolic x coords={explicit (SR),presupposed (PR),open-ended (IPA)},
  xtick=data, x tick label style={font=\scriptsize},
  legend style={at={(0.5,1.03)}, anchor=south, font=\scriptsize,
                draw=none, fill=none, legend columns=2,
                /tikz/every even column/.append style={column sep=6pt}},
  every axis plot/.append style={draw=none},
  ymajorgrids, major grid style={gray!16},
  nodes near coords={\pgfmathprintnumber[fixed, fixed zerofill,
    precision=2]{\pgfplotspointmeta}},
  every node near coord/.append style={font=\scriptsize},
]
\addplot[fill=gray!55,
  every node near coord/.append style={xshift=-3pt}]
  coordinates {(explicit (SR),.81)(presupposed (PR),.44)(open-ended (IPA),.06)};
\addplot[fill=figmodelb,
  every node near coord/.append style={xshift=3pt}]
  coordinates {(explicit (SR),.63)(presupposed (PR),.69)(open-ended (IPA),.13)};
\legend{gpt-4o-mini,gpt-4o}
\end{axis}
\end{tikzpicture}
\caption{Draft-anchored verification blindness: stale-premise recall
is high or mixed on explicit and presupposed probes but collapses on
open-ended IPA probes, where the stale state is unstated. The pattern persists at
frontier scale (gpt-5.5: $0.38$; \S\ref{sec:results}) and worsens under
few-shot or chain-of-thought prompting ($0.38\to0.13$).}
\label{fig:blindness}
\end{figure}

\textbf{(2) Method.} We introduce \emph{provenance-verified transition
assembly} (VTA), where ``verified'' refers to provenance and chronology
rather than semantic correctness: bidirectional lifecycle auditing,
lifecycle-focused evidence expansion, candidate transition proposal,
deterministic entry-level validation, and verified-only typed repair.
\textbf{(3) Controlled evidence with explicit boundaries.} Under the
strict single-query protocol VTA gains $+\StrictDelta$ points paired over
the locked predecessor (\StrictMacro{} vs.\ \StrictBaseMacro{}), IPA-led,
reproduced by the benchmark's disjoint judge, and separated from
budget/evidence confounds by a full-scale matched control; a
privileged scenario-joint variant (\OffMacro{}) is reported only as an
upper bound. All records are released (evidence pairs, causal paths,
validation decisions, costs), together with the boundaries: draft
regressions, same-family judge risk, post-development full-set use, a
failed 4B distillation, and a harder authored set with no gain; no number
here implies deployment safety.

\section{Related Work}

\paragraph{Draft revision against world knowledge.}
Post-hoc decompose--verify--revise pipelines are well established,
including RARR \citep{gao2023rarr}, CoVe \citep{dhuliawala2024cove}, and
Verify-and-Edit \citep{zhao2023verifyedit}; subsequent work compresses
the loop into smaller trained editors \citep{chen2023purr,mishra2024fava}
or fine-tuned in-domain verifiers \citep{pave2026,finground2026}. These
methods verify claims against evidence treated as temporally stable. They
do not model evidence lifecycles or support a user-verification action,
and they typically repair text locally rather than regenerate downstream
behavior. We further show (\S\ref{sec:results}) that their common
draft-anchored formulation is insufficient for implicit policy
adaptation.

\paragraph{Small trained verifiers.}
MiniCheck trains sub-1B--7B fact checkers on synthetic claim--document
data at a fraction of frontier cost \citep{tang2024minicheck}. We test
whether the recipe transfers to state semantics (lifecycle verdicts over
evolving personal memory); our 4B instantiation, trained with
construction-derived labels, does not yet match the frontier auditor
(supplement).

\paragraph{Persona and profile consistency.}
A pre-LLM lineage audits responses against a stored profile. Dialogue
NLI formulates response--persona consistency as entailment
\citep{welleck2019dnli}; Generate-Delete-Rewrite removes
persona-inconsistent tokens from a response prototype \citep{song2020gdr};
and KvPI classifies response--profile relations as entailed,
contradicted, or irrelevant \citep{song2020kvpi}.
The closest prior work in structure is Post Persona Alignment (PPA)
\citep{ppa2025}, which also intervenes after generation. We retain this
stage but change the verification target and the response action: PPA
aligns a response with a static persona set, whereas our lifecycle
verdicts distinguish currently valid, superseded, and unresolved user
states --- a superseded premise carries the attribute's current value, and an
unresolved one can trigger a user-confirmation request, distinctions an
entailment/contradiction scheme cannot represent. PrefEval's
prompt-based self-critique checks drafts against stated preferences
\citep{prefeval2025}. Neither PPA nor PrefEval models temporal
invalidation of previously valid user states.

\paragraph{Agent memory governance.}
Memory governance operates primarily at write time or read time.
Write-time methods address consolidation and conflict resolution
\citep{mem0,memgpt}, temporal edge invalidation \citep{zep}, learned
write policies \citep{memoryr1,memalpha}, supersession training
\citep{supersede2026}, and deterministic freshness resolution
\citep{freshness2026}; read-time methods filter retrieval or construct
temporal evidence graphs \citep{trace2026}. CUPMem, the \stale{}
authors' prototype, adjudicates updates at write time and constrains the
readout so that generation follows prior adjudication \citep{stale2026}.
Verification-gated persona transitions \citep{personaverif2026} validate
symbolic traces before committing updates to the \emph{agent's} persona;
in contrast, we audit responses for dependence on the \emph{user's}
evolving state and regenerate behavior rather than approve a memory
update. Memora \citep{memora2026}, MemConflict \citep{memconflict2026},
EvolveMem \citep{evolvemem2026}, and MemTrace \citep{memtrace2026} (the
last closest to our diagnosis: reachable evidence often goes
unused) diagnose or optimize memory conflicts but do not provide the
response-level repair studied here; over-personalization benchmarks
\citep{opbench2026} target the verify-question noise our repair-only point
avoids.
User as Code \citep{useracode2026} compiles user state into executable
memory, strengthening the information supplied to generation; our
response-side intervention is complementary, since even a visible update
may be ignored. Concurrent work studies state roles, learned transition verification,
and bounded topology repair
\citep{atma2026,trustmem2026,ariadnemem2026,allmem2026}; the general
abstraction is not our claim. We focus on a post-generation blind
spot --- responses may depend on stale user states without stating them
--- and verify each proposed transition's entry-level provenance and
chronology before the transition can authorize a typed behavioral
repair.

\section{Problem Setup}
\label{sec:setup}

An assistant serves a user whose state evolves over time. Its memory is a
store $M = \{(t_i, e_i)\}$ of timestamped entries (dialogue sessions or
extracted facts), ordered oldest-first. A later entry may implicitly
invalidate an earlier one, as in the implicit-conflict regime of \stale{}
\citep{stale2026}. Each conflict scenario contains an old state
$m_{\text{old}}$, an updated state $m_{\text{new}}$, and three probes.
\textbf{State resolution (SR)} asks whether the system recognizes that
$m_{\text{old}}$ no longer holds. \textbf{Premise resistance (PR)} tests
whether it rejects a query that presupposes $m_{\text{old}}$.
\textbf{Implicit policy adaptation (IPA)} tests whether an open-ended
plan or recommendation reflects $m_{\text{new}}$ without explicit
prompting.

\paragraph{The audit as a dependency check.}
Let $\sigma$ map attributes to the user's current values. A draft $d$
depends on attribute $a$ if changing $\sigma(a)$ would change the
substance of $d$, whether or not $d$ mentions $a$; call this attribute
set $\mathrm{Dep}(d)$. Draft-anchored verification approximates it by
$\mathrm{Dep}_{\text{say}}(d)$,
the premises stated in $d$; for open-ended queries
$\mathrm{Dep}_{\text{say}}(d) \subsetneq \mathrm{Dep}(d)$, and the
omitted dependencies often concern changed attributes. The
state-anchored pass therefore enumerates the changed set $C$ and tests
each member directly, targeting $\mathrm{Dep}(d) \cap C$.
$\mathrm{Dep}(d)$ is a counterfactual definition, not a direct
observable: at runtime an LLM consistency judgment \emph{estimates}
membership rather than performing the state intervention, and retrieval
misses make the enumerated $C$ itself incomplete. Given $M$, a query $q$, and a draft $d \sim G(M, q)$ from the
memory-augmented generator $G$, the audit produces
$P = \{(p_j, v_j, c_j, b_j)\}$: premise, verdict
$v_j \in \{\valid, \staleflag, \unknown\}$, current value when
available, and materiality. The adapter returns $d' = A(M, q, d, P)$
under the typed directives of \S\ref{sec:directives}; if no directive
fires, $d' = d$.

\paragraph{Memory settings.}
We evaluate two memory settings. In the \emph{oracle-memory
development setting}, $M$ contains only the scenario-relevant sessions,
isolating the response-side auditing stage, where CUPMem's IPA
failures occur; all
such analyses are labeled development results. The \emph{official
full-history protocol} is closer to deployment: all 50 chronological
sessions ($\sim$150K tokens), no oracle annotations at runtime, one
independently generated response per query, and \stale{}'s released
judge scoring the three responses jointly once generation is complete
(oracle
indices are used only for offline retrieval-recall analysis). The locked
predecessor follows this independent-query scope; we evaluate VTA both
under it (\emph{strict single-query VTA}) and with a scenario-level
extension jointly auditing all three question--draft pairs
(\emph{scenario-joint VTA}).

\section{\system}

\begin{figure*}[t]
\centering
\includegraphics[width=.8\textwidth]{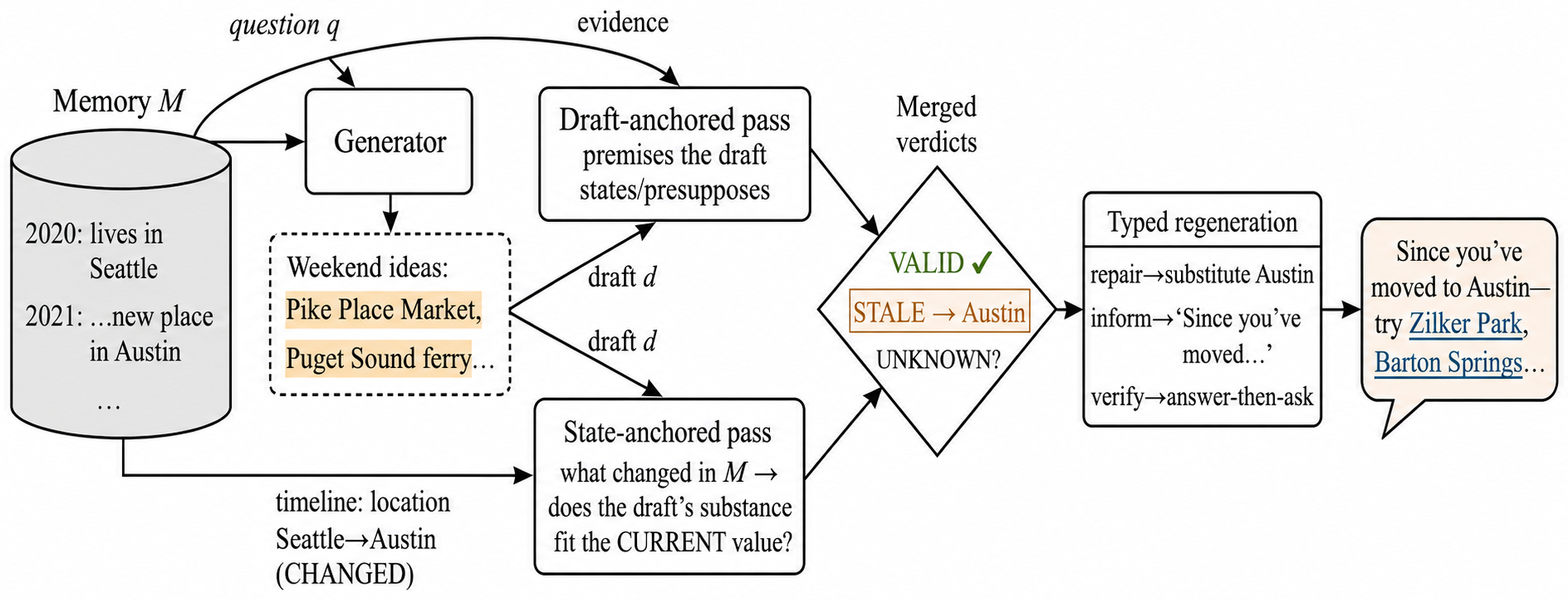}
\caption{\system{} on a worked example: the weekend-ideas draft still
leans on Seattle after the move to Austin. The draft-anchored pass
adjudicates premises the draft states or presupposes; the state-anchored
pass checks the draft's \emph{substance} against each changed
attribute's current value. Merged verdicts drive typed regeneration:
\emph{repair}, \emph{correct-and-inform}, or \emph{verify} (answer, then
ask). A provenance- and chronology-verified transition (\S\ref{sec:vta})
must authorize the repair; otherwise the base verdicts fail open. Under
the strict protocol each probe runs this pipeline independently.}
\label{fig:pipeline}
\end{figure*}

\system{} uses a fixed, role-structured audit pipeline
(Fig.~\ref{fig:pipeline}): the \textbf{Evidence} role retrieves dated
statements (\S\ref{sec:retrieval}), the \textbf{Dependency} role
extracts the premises the draft relies on (\S\ref{sec:draftpass}), the
\textbf{Temporal} role constructs the user's state timeline
(\S\ref{sec:statepass}), and an optional \textbf{Skeptic} challenge pass exists but is disabled in
every evaluated configuration (supplement). VTA then proposes and validates candidate state transitions
(\S\ref{sec:vta}), after which the \textbf{Arbiter} produces
deterministic typed directives for the \textbf{Adapter}
(\S\ref{sec:directives}). Every inter-role message is a typed verdict
with evidence; control flow does not depend on model-authored plans, and
\S\ref{sec:official} reports each role's failure modes separately.

\subsection{Evidence Role: Query-Time Retrieval over Full Histories}
\label{sec:retrieval}
Under the official protocol, the audit can adjudicate only the evidence
returned by retrieval, and superseding evidence is difficult to
retrieve: it often appears as a life-event disclosure whose wording
overlaps with neither the question nor the invalidated assertion (e.g.,
a new-lease remark implying a move without naming either city). We
index each session's user turns as overlapping sentence windows and
select evidence in four deterministic steps, each ablated on the frozen
dev slice. We (i)~rank question-anchored lexical matches with
per-session deduplication and temporal diversity; (ii)~reserve
change-scan slots for high-scoring, query-independent update
disclosures; (iii)~add each selected session's strongest disclosure and
best question-matching window; and (iv)~search strictly earlier sessions
for antecedents of each disclosure, pairing the update with the
assertion it supersedes. On the frozen dev slice this procedure increases strict window-level
recall of the old/new \emph{pair} from $.125$ to $.708$ with a
$\sim$5.7K-token budget and no retrieval-time LLM calls ($.580$ pooled
over all 400 official scenarios; per-dimension breakdown in the
supplement). The generator always receives this
compact readout. Scenario-joint VTA additionally expands
runtime-selected sessions (never oracle-selected) and
reranks their user-side windows using all three probes and generic
lifecycle cues. This expanded evidence is available only to the audit,
not to the generator or to an independently processed probe.

\subsection{Dependency Role: Draft-Anchored Premise Decomposition}
\label{sec:draftpass}
Given an oldest-to-newest readout $M$, question $q$, and draft $d$,
the Dependency role extracts atomic premises about the user's state ---
explicit statements and implicit dependencies, such as a presupposition
in $q$ the draft accepts. Each premise is evaluated with $M$ as the sole
evidence source: \valid{} if the most recent relevant evidence supports
it, \staleflag{} if a later entry changes or invalidates it (the
current value is recorded), and \unknown{} if $M$ neither confirms nor
refutes it; a materiality bit marks whether the answer meaningfully
depends on the premise.

\subsection{Temporal Role: State-Anchored Timeline Consistency}
\label{sec:statepass}
Draft-anchored decomposition cannot surface premises the draft never
states. The state-anchored pass reverses the direction: it extracts an
attribute timeline from $M$ (current value, previous value, change
status) and checks whether the draft's substance is consistent with
each current value --- \staleflag{} when it relies on the previous
value or materially ignores the update, \valid{} when consistent, and
\unknown{} when the required attribute is not established by memory.

\subsection{Provenance-Verified Transition Assembly}
\label{sec:vta}
The three probes in a scenario share one history and one annotated
transition, yet independent audits may resolve SR correctly while
accepting the stale premise in PR. The transition proposer reads
question--draft pairs and outputs an attribute, old and current values,
two evidence quotations, a short causal path, and the affected response
dimensions. It runs under two information scopes: \emph{single-query},
where each call sees only one probe (the strict, protocol-matched
setting), and \emph{scenario-joint}, where one call observes all three
question--draft pairs. Candidate generation prioritizes recall and may
include ordinary causal or affordance relations, such as
move$\rightarrow$possession; only memory entries can serve as evidence
--- questions and drafts cannot.

Authorization is deterministic. Each quotation must match at least
80\% of its content tokens within a single rendered memory entry. The
validator uses the matched entry's timestamp rather than a
model-generated date and requires $t_{\rm new}>t_{\rm old}$, both state
values, a material application, and a causal path with at least two
nodes. Candidates with reversed chronology, cross-entry quotations,
ungrounded evidence, or parse failures cannot authorize a \staleflag{}
verdict. This design combines deterministic freshness aggregation
\citep{freshness2026} with the question-conditioned causal reasoning
required by implicit (T2) conflicts.

\subsection{Arbiter and Adapter: Deterministic Typed Regeneration}
\label{sec:directives}
The arbiter combines both audit passes with validated VTA paths. When
a dimension has a verified transition, only that path can authorize a
repair; unverified verdicts remain in telemetry but cannot override the
validated chronology. Without a verified transition the system falls
back to the base audit's verdicts, whose material \staleflag{}/\unknown{}
findings can still fire repair or verify directives --- ``verified-only''
constrains the transition channel, not the base audit. Directives are derived deterministically from
the final verdicts: material \staleflag{} triggers \emph{repair} and
\emph{correct-and-inform} (a one-sentence acknowledgment), and material
\unknown{} triggers \emph{verify} (answer, then ask). We use
deterministic rules because model-generated directives frequently asked
for confirmation even after a stale state was established. If no directive fires, the draft passes through; if the audit output
cannot be parsed, the system returns the unmodified draft (rates in
\S\ref{sec:overhead}).

\subsection{Fine-Tuned Auditor: Synthetic Data with Labels by Construction}
\label{sec:synth}
We train a Qwen3-4B auditor on synthetic scenarios from our own
taxonomy (30 attribute types $\times$ 20 domains $\times$ 5 change
patterns $\times$ 4 draft modes): a structured JSON scenario (old and
new values, dates, distractors) is sampled first, dialogue snippets
render the transition without explicit negation, and a probe and a
mode-controlled draft follow. Gold premises, verdicts, current values,
and materiality are derived from the scenario structure, not from a
model's judgment of the rendered text; 35\% of drafts are correct by
construction to supervise over-correction control. A contamination
audit run before training (exact 8-gram and embedding similarity against all
four evaluation sets) reports no collisions and no flagged items.

\section{Experimental Setup}
\label{sec:exp}

\paragraph{Benchmarks.}
All benchmarks are evaluation-only. (1) \stale{} \citep{stale2026}
provides 400 expert-validated implicit-conflict scenarios, each with
three probes (SR/PR/IPA). (2) From LongMemEval \citep{longmemeval}, we use knowledge-update and
\texttt{\_abs} abstention questions (the correct answer to the latter
is to decline --- a natural probe of excessive verification). (3) From
MemoryAgentBench FactConsolidation \citep{mab2025}, we use
counterfactual fact stores where the latest edit must override earlier
facts and parametric knowledge (single-/multi-hop, four context
lengths). (4) From LoCoMo \citep{locomo}, we
use ordinary conversational QA (categories 1, 2, and 4) as an
over-correction probe; because these examples may contain genuine
updates, the \staleflag{} rate is only an upper bound on the
false-invalidation rate (\S\ref{sec:overhead}). LoCoMo and MemoryAgentBench are
adapted into observation-fact and fact-list probes rather than their
original incremental protocols (prospective freezes in the supplement).
(5) \horizonbench{}~\citep{horizonbench2026}: 120 evolved-preference
items (61 users; authoring models distinct from our gpt-5.5 stack), full
draft$\to$audit$\to$repair pipeline over a \emph{gold-derived} store
rendering each event's \texttt{from}$\to$\texttt{to} change; a gpt-5.5
option-mapping judge (checked by a disjoint Gemini re-judge) scores which
preference each response expresses. (6) A frozen 120-scenario
Claude-authored hard lifecycle set (six stress categories, protocol
committed before generation; deterministic keys, no judge). Strict
\stale{} and both external sets ran under pre-fixed configurations; the
scenario-joint variant, ablations, matched controls, and repair-only
selection are post-development analyses, labeled as such.
Development used frozen slices of at most 50 items (seed 42; committed
ID lists); no prompt was tuned on any full test set.

\paragraph{Systems.}
In the development setting, all methods use the same generator
(gpt-5.4-mini), memory readout, and benchmark-specific judge prompts;
the official-protocol stack differs and is specified with its results
(\S\ref{sec:official}). Baselines include a no-audit condition; a CUPMem-style
adjudicated-store summary inserted into the generator context, which
tests whether exposing a resolved state to the generator suffices;
prompt-based self-critique \citep{prefeval2025}; factored CoVe
\citep{dhuliawala2024cove}; RARR applied to the memory store
\citep{gao2023rarr}; and timeline-conditioned regeneration without
verdicts or directives. We evaluate four \system{} variants:
draft-anchored only, state-anchored only, bidirectional, and
bidirectional with the fine-tuned 4B auditor. All API calls are cached
and their costs are logged; every run records its configuration, git
hash, and seed.

\paragraph{Implementation details.}
In the official-protocol runs, Claude Sonnet~5 serves as generator
and fallback auditor (five audit calls per query; the sixth call in
Table~\ref{tab:config} is the transition proposal, and the adapter fires
only on a directive),
Gemini~3.1 Flash-Lite as the scenario-joint VTA proposer/adapter, and
gpt-5.5 as proposer, adapter, and judge in the strict confirmation runs.
Decoding is greedy where supported; the Anthropic API samples without
a seed at
$T{=}1.0$ and neither Gemini nor the gateway exposes a seed, so each arm
is one sampled output per item unless identified as a replicate. The
artifact includes per-item records, manifests, and scripts that
reproduce all reported metrics from cached responses (cache and 4B
checkpoint, 674\,MB combined, with the camera-ready version). We fine-tune Qwen3-4B with
QLoRA on 599 synthetic examples using a single 8\,GB GPU
(\S\ref{sec:synth}).
The predecessor was developed only on frozen development slices; VTA's
post-development full-set use is discussed in Limitations, and a
construction-frozen synthetic set (\S\ref{sec:official}) provides
limited mechanism-generalization evidence in place of an untouched
\stale{} split.

\paragraph{Blind human protocol.}
Two annotators independently labeled the same 100 frozen cells
(scenario--probe judgment units), with
source, phase, scenario ID, and automated verdicts hidden. The initial
blind draft/predecessor package predated VTA; before inspecting any VTA
labels, we froze an extension containing VTA outputs for the same cells.
We report each annotator's paired exact McNemar test; neither annotator used
\textsc{unsure}. The released artifact includes the protocol, file
hashes, annotation sheets, and analysis code.

\paragraph{Judge validity.}
The scenario-joint score uses \stale{}'s released Gemini judge and
rubric; the VTA proposer and adapter share that family, whereas the
generator and fallback auditor use Claude. A deterministic local
Qwen3-4B re-judge of all $n{=}\OffLocalJudgeN$ scenarios reproduces the
VTA/predecessor ordering
(\OffLocalPriorVta{}/\OffLocalPriorBase{};
$\OffLocalPriorWins{}:\OffLocalPriorLosses{}$,
$p<10^{-\OffLocalPriorPExp}$), and both blind annotators reproduce it
on the frozen human cells (\HumanImproveA{}:\HumanRegressA{} and
\HumanImproveB{}:\HumanRegressB{}, both significant; agreement
$\kappa{=}\HumanKappa$). These checks validate the ranking, not the absolute \OffMacro{}
estimate (supplement); strict-arm judge validation is in
\S\ref{sec:official}.

\section{Results}
\label{sec:results}

\begin{table*}[t]
\centering\small
\setlength{\tabcolsep}{6pt}
\begin{tabular}{l cc cc cc l}
\toprule
& \multicolumn{2}{c}{SR} & \multicolumn{2}{c}{PR} & \multicolumn{2}{c}{IPA} & \\
\cmidrule(lr){2-3}\cmidrule(lr){4-5}\cmidrule(lr){6-7}
System & T1 & T2 & T1 & T2 & T1 & T2 & Macro [95\% CI] \\
\midrule
\multicolumn{8}{l}{\emph{Independent-query strict protocol (headline; gpt-5.5 judge; frozen drafts/base audits)}}\\
retrieval only (same frozen drafts) & .615 & .290 & .215 & .125 & .635 & .490 & .395 [.367,\,.424] \\
\;+ predecessor (same drafts) & .795 & .675 & .745 & .575 & .725 & .600 & .686 [.653,\,.718] \\
\;+ \textbf{single-query VTA (strict)} & \textbf{.800} & .630 & \textbf{.805} & \textbf{.620} & \textbf{.805} & \textbf{.755} & \textbf{.736} [.707,\,.767] \\
\midrule
\multicolumn{8}{l}{\emph{Scenario-joint upper bound (privileged; released Gemini judge)}}\\
retrieval only & .605 & .295 & .210 & .130 & .595 & .415 & .375 [.344,\,.407] \\
\;+ predecessor (bidirectional ensemble) & .795 & .675 & .755 & .595 & .705 & .555 & .680 [.642,\,.715] \\
\;+ scenario-joint VTA$^\dagger$ & \textbf{.940} & \textbf{.850} & \textbf{.935} & \textbf{.870} & \textbf{.890} & \textbf{.790} & \textbf{.879} [.853,\,.904] \\
\bottomrule
\end{tabular}
\caption{\stale{} full-history protocol (all 400 scenarios (T1/T2 $=200/200$); 50-session
histories; no runtime oracle annotations). Upper panel: strict
independent-query confirmation --- frozen drafts/base audits, one locked
gpt-5.5 judge; protocol-matched within the panel. Lower panel:
scenario-joint VTA$^\dagger$ reads all three probes jointly, a
privileged upper bound not comparable to independent-query systems.
$^\dagger$one transition call over all SR/PR/IPA pairs plus expanded
evidence, provenance/chronology-validated. CI: stratified scenario
bootstrap; T1/T2 = explicit/implicit conflicts.}
\label{tab:official}
\end{table*}

\begin{table}[t]\centering\footnotesize
\setlength{\tabcolsep}{2.6pt}
\begin{tabular}{l l l c l}
\toprule
System & Evidence & P/A/J & Calls & Authority \\
\midrule
retrieval only & per-query & --/--/Ge & 0 & --- \\
predecessor & per-query & --/Cl/Ge & 5 & directives \\
joint VTA & 3-probe union & Ge/Ge/Ge & 5.33 & verif.-only \\
strict VTA & per-query+exp. & 55/55/55 & 6 & verif.-only \\
\bottomrule
\end{tabular}
\caption{System configurations. P/A/J = proposer/adapter/judge; Cl/Ge/55
= Claude, Gemini 3.1 Flash-Lite, gpt-5.5. All rows share the Claude
generator; VTA rows reuse the frozen base audits; ``exp.'' = audit-side
evidence expansion.}
\label{tab:config}
\end{table}

\subsection{Full-History \stale{} Results}
\label{sec:official}
Table~\ref{tab:official} compares retrieval alone, the locked
predecessor --- \system{}'s bidirectional ensemble, frozen before VTA
existed --- and scenario-joint VTA on the same 400 scenarios
(configurations in Table~\ref{tab:config}); all three
use the same compact generator readout, Claude generator, and released
joint judge. Their macro scores are \OffDraftMacro{}, \OffPriorMacro{},
and \OffMacro{}~$[\OffCILow,\OffCIHigh]$, respectively.

Relative to its own drafts scenario-joint VTA fixes \OffFix{} cells and
breaks \OffBreak{} (cluster-robust scenario sign test
\OffScenImproved{}:\OffScenRegressed{}, $p<10^{-\OffPExp}$). But it reads
all three probes jointly while the target runner processes them
independently, so the $.879$ is a scope-confounded upper bound, not a
protocol-matched result (details and the $306/979$ provenance-valid pairs:
supplement). The strict independent-query confirmation below is the
comparable number.

\paragraph{Strict independent-query confirmation.}
Under this protocol, strict single-query VTA reuses the predecessor's frozen drafts
and base audits; each proposer sees only its own query, draft, and
audit evidence; and the judge scores the three finished responses
afterward. Proposer, adapter, and judge were locked to gpt-5.5 before
execution, with the predecessor re-judged by the same judge
(Table~\ref{tab:official}, headline panel). Strict VTA processes
$\StrictN$ of 400 scenarios without a parse failure and scores
\StrictMacro{}~$[\StrictCILow,\StrictCIHigh]$ against
\StrictBaseMacro{}: $+\StrictDelta$ points paired (95\% CI
$[+\StrictDeltaLow,+\StrictDeltaHigh]$), \StrictScenWins{} scenarios
better and \StrictScenLosses{} worse ($p<10^{-\StrictScenPExp}$;
Fig.~\ref{fig:gains}). The released Gemini judge, disjoint from the gpt-5.5 stack,
reproduces the gain ($.738$ vs.\ $.680$; $92{:}40$,
$p{=}7\times10^{-6}$), and it is stable across runs: \VarNRuns{}
independent full-400 draws score \VarMean{}${\pm}$\VarStd{}, the lowest
still $+\VarMinAdv{}$ over the predecessor. IPA moves
$\StrictIpaWins{}:\StrictIpaLosses{}$ (T2
\StrictTTwoIpaWins{}:\StrictTTwoIpaLosses{}), PR
\StrictPrWins{}:\StrictPrLosses{}, and SR stays flat
(\StrictSrWins{}:\StrictSrLosses{}, with a T2 regression pocket of
\StrictTTwoSrWins{}:\StrictTTwoSrLosses{} we return to below); proposer
isolation authorizes fewer transitions at comparable gold-pair alignment
(supplement).

\paragraph{Generalization: the full pipeline on a second benchmark.}
On the \horizonbench{} adaptation (protocol in \S\ref{sec:exp}),
auditing the open-ended draft lifts current-preference accuracy
\HbFullBase{}$\to$\HbFullVta{} and lowers reversion to the superseded
value \HbFullStaleBase{}$\to$\HbFullStaleVta{}
(\HbFullWins{}:\HbFullLosses{}, $p{=}\HbFullP$; user-clustered
$p{=}\HbFullClustP$; all three authoring families; the disjoint re-judge
agrees). But a matched no-transition control (same evidence, adapter, and
budget) already reaches \HbCtrlNotrans{}
(\HbCtrlNotransBaseWins{}:\HbCtrlNotransBaseLosses{} over base): unlike
\stale{}, most of this external gain is the draft-side audit itself, and
the transition machinery adds only a non-significant increment
(\HbCtrlVtaNotransWins{}:\HbCtrlVtaNotransLosses{},
$p{=}\HbCtrlVtaNotransP$) --- expected, since the gold-derived store
already encodes the change (MCQ probe and its reasoning-matched control:
supplement). A same-family construct set matches direction
(Fig.~\ref{fig:gains}); a harder authored set bounds the gain
(\S\ref{sec:limitations}).

\definecolor{figimproved}{HTML}{2563EB}
\definecolor{figregressed}{HTML}{E8590C}
\begin{figure}[t]
\centering
\begin{tikzpicture}
\begin{axis}[
  width=.92\columnwidth, height=3.0cm,
  xbar stacked, xmin=-55, xmax=112,
  ytick={0,1,2,3},
  yticklabels={
    {synthetic $n{=}80$},
    {replicate $n{=}100$},
    {disjoint judge},
    {strict (headline)}},
  y tick label style={font=\scriptsize, align=right},
  ytick style={draw=none},
  axis line style={gray!50},
  xlabel={scenarios regressed $\leftarrow$ $|$ $\rightarrow$ improved},
  xlabel near ticks, x label style={font=\scriptsize},
  xtick={-40,0,40,80}, xticklabels={40,0,40,80},
  x tick label style={font=\scriptsize, gray!40!black},
  bar width=9pt, enlarge y limits=0.16,
  every axis plot/.append style={draw=white, line width=0.5pt},
  xmajorgrids, major grid style={gray!14},
  clip=false,
]
\addplot[fill=figregressed] coordinates {(0,0) (-9,1) (-40,2) (-37,3)};
\addplot[fill=figimproved] coordinates {(5,0) (21,1) (92,2) (84,3)};
\draw[gray!55, line width=0.5pt] (axis cs:0,-0.55) -- (axis cs:0,3.55);
\node[font=\scriptsize, anchor=west] at (axis cs:7,0) {5};
\node[font=\scriptsize, anchor=west] at (axis cs:23,1) {21};
\node[font=\scriptsize, anchor=west] at (axis cs:94,2) {92};
\node[font=\scriptsize, anchor=west] at (axis cs:86,3) {84};
\node[font=\scriptsize, gray!35!black, anchor=east] at (axis cs:-2,0) {0};
\node[font=\scriptsize, gray!35!black, anchor=east] at (axis cs:-11,1) {9};
\node[font=\scriptsize, gray!35!black, anchor=east] at (axis cs:-42,2) {40};
\node[font=\scriptsize, gray!35!black, anchor=east] at (axis cs:-39,3) {37};
\end{axis}
\end{tikzpicture}
\caption{Paired scenario wins (blue, right) and losses (orange, left)
for strict VTA over the locked predecessor; ties omitted. Same judge:
$84{:}37$, $p{<}10^{-4}$, IPA $53{:}6$. Disjoint Gemini judge:
$92{:}40$, $p{=}7{\times}10^{-6}$, IPA $68{:}17$. Replicate
($n{=}100$): $21{:}9$, $p{=}.043$. Synthetic set ($n{=}80$): $5{:}0$,
$p{=}.06$.}
\label{fig:gains}
\end{figure}
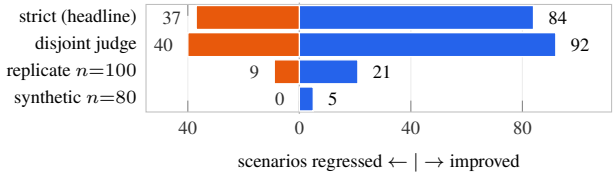

\paragraph{Where the joint advantage lives.}
A locked factorial ($n{=}100$) adds one cross-probe
channel at a time: sibling questions ($.727$) or
drafts ($.730$) do not improve over strict ($.733$), whereas the joint
evidence union drops to $.683$ ($p{=}.004$) --- the scenario-joint
advantage requires the fully joint shared-graph call (supplement).

\paragraph{Which components are associated with the gain.}
A second locked factorial ablates one VTA component at a time (same
holdout/stack). Removing transition proposal and validation drops accuracy
$.733{\to}.677$ ($p{=}.0013$); skipping validation gives $.740$ ($298/300$
outputs unchanged) and disabling verified-only authority $.733$ (verdicts
unchanged), so accuracy comes from transition proposal over expanded
lifecycle evidence and typed repair, not the gate. The gate barely moves
verdicts here because the proposer rarely emits a bad candidate ($234$ of
$237$ safety-suite hard negatives abstained; supplement); its value is
auditability and a structural bound on weaker proposers, quantified by
injection: the gate accepts every valid candidate, rejects all
\GateRejectStruct{}/\GateStructTotal{} chronology/grounding/path
violations, and passes \GateNonsupVerify{}/\GateNonsupN{} semantically
non-superseding ones --- provenance and chronology, not semantic
supersession (supplement). Where authorizations align with the annotated
pair, accuracy is near-perfect ($.997$, $55{:}0$; supplement). This holds at full scale: a matched control
on all 400 scenarios (same evidence, adapter, and call budget,
transition machinery removed) scores \DirectMacro{}, only
$+\DirectDelta$ over the predecessor and not significant
($\DirectWins{}{:}\DirectLosses{}$); strict VTA beats it by
$+\TransDelta$ points ($\TransWins{}{:}\TransLosses{}$,
$p<10^{-\TransPExp}$, all dimensions; the disjoint Gemini judge agrees,
$+\TransDeltaDisj{}$, $\TransWinsDisj{}{:}\TransLossesDisj{}$). The
budget, evidence, and adapter recover almost none of the gain.

In the joint run the \OffFailedCells{} failed cells split across
retrieval, no-verified-path, off-target, and adapter/judge causes
(supplement). Because its proposer/adapter shares a family with the judge,
we treat the $.879$ joint score only as an upper bound, not
independent-judge evidence.

\subsection{Development Setting: The Audit Recovers PR and Lifts IPA}
In the oracle-memory development setting (400 examples per
dimension), bidirectional \system{} --- later locked as the
predecessor --- raises overall accuracy from 28.3\% to 67.3\%
(PR 145:0, IPA 135:9; both $p<10^{-3}$).
\textbf{(a)} PR survives an in-harness CUPMem-analogue matched control
(supplement). \textbf{(b)} The state-anchored
pass improves IPA by 31.5 points (135 corrected vs.\ 9 regressed of 400);
draft-anchored auditing tracks no-audit. \textbf{(c)} \system{} beats every
development baseline (CUPMem analogue $p{=}.012$; CoVe, RARR-on-store, and
self-critique all $p<.05$; supplement).

\subsection{Why: The Bottleneck Is Extraction, Not Verification}
We measure intrinsic auditor quality using stale-premise recall:
whether some \staleflag{} premise matches $m_{\text{old}}$. For a frontier
gpt-5.5 draft-anchored auditor, recall is high on SR ($.88$) and PR
($1.0$) but falls to $.38$ on IPA, where the stale dependency is
unstated. Few-shot and chain-of-thought prompting reduce rather than
recover it ($.38 \to .13$), and scale does not remove the gap. At mid scale the state-anchored pass roughly doubles IPA recall. The gain is not just prompt augmentation: on the fixed 100-scenario
holdout (identical evidence, generator, judge), the audit ladder rises
from timeline-conditioning ($.517$ from a $.373$ base) to the locked
ensemble $.680$, $16.3$ points above augmentation (supplement). Nor is it Claude-specific: a Gemini auditor/adapter
under a disjoint gpt-5.5 judge lifts its drafts
\OffBackboneDraft{}$\to$\OffBackboneFinal{}
($p<10^{-\OffBackbonePExp}$). \textbf{Lifecycle verdicts and typed repairs matter}: NLI labels in
place of \valid{}/\staleflag{}/\unknown{} drop base-audit accuracy
$.640\to.537$, and an untyped consistency rewrite drops it to $.503$
(both $p<10^{-4}$; supplement): ``contradicted'' names no current value
to repair and no unresolved state to verify.

\subsection{Over-Correction and Overhead: Measured and Mitigated}
\label{sec:overhead}
\paragraph{Over-correction.}
An auditor that helps on conflicts can hurt elsewhere: the
mid-capability auditor (gpt-4o-mini; supplement) sharply lowers
conflict-free accuracy
(LongMemEval knowledge-update $.816\to.579$; LoCoMo $.600\to.433$); the
frontier auditor holds aggregate accuracy but still flags $.367$ of
ordinary LoCoMo questions, and the predecessor's LongMemEval transfer
probe ($n{=}72$, disjoint judge) slips from $.861$ to $.722$ ($5{:}15$).
A construction-labeled safety suite (seven regimes, 277 items;
supplement) puts numbers on the boundary: the predecessor reaches
$.975$ true-change recall but $.05$--$.58$ false invalidation by
regime. Strict VTA, run end to end on the same suite, cuts false
material-\staleflag{} to $.00$--$.05$ (repair-class fires on 7 of 237
conflict-free items) and still catches $.875$ of true changes. The residual is \emph{verify} question noise: a gpt-5.5 quality judge
rates the \VerifyNRewrites{} verify-only rewrites
\VerifyPreservedN{}/\VerifyWeakenedN{}/\VerifyChangedN{}
preserved/weakened/changed, $\VerifyRelevantRate\%$ on-topic but
$56$--$\VerifyUnnecessaryRate\%$ unnecessary and $10$--$42\%$ intrusive
(three judge families agree; changed cells bound the harm at
\VerifyTotalToWrong{}/\VerifyTotalFixed{} correct$\to$wrong/reverse;
supplement). When we compose
each policy's three actual responses and re-judge them end to end ---
without score splicing --- the post-hoc-selected \emph{repair-only} policy lands at
\RepairOnlyMacro{} (splice predicted $.726$), still beats the baseline
($\RepairOnlyWins{}{:}\RepairOnlyLosses{}$, $p{=}\RepairOnlyP$;
disjoint Gemini agrees, supplement), and rewrites $3\%$ of
conflict-free examples vs.\ $80\%$. It \emph{deepens} the T2-SR pocket
(\StrictTTwoSrWins{}:\StrictTTwoSrLosses{} $\to$
$\RepairOnlyTTwoSrWins{}{:}\RepairOnlyTTwoSrLosses{}$): the verify
gain there is diffuse acknowledgment, and gating removes it with the
noise. Fig.~\ref{fig:pareto} shows both operating points; we recommend
repair-only as the deployment default wherever unnecessary personalized
follow-up questions are costly, treating full VTA as the
accuracy-oriented point. Safety evidence is
from conflict-focused suites (LoCoMo lacks gold conflict labels, so
invalidation is an upper bound); strict was not run on LongMemEval or
natural traffic (supplement).
\definecolor{figrepaironly}{HTML}{2563EB}
\definecolor{figfullvta}{HTML}{E8590C}
\begin{figure}[t]
\centering
\begin{tikzpicture}
\begin{axis}[
  width=.95\columnwidth, height=2.5cm,
  xlabel={conflict-free rewrite-trigger rate (\%)},
  ylabel={\stale{} macro},
  x label style={font=\scriptsize}, y label style={font=\scriptsize},
  xmin=-0.06, xmax=1.02, ymin=.668, ymax=.762,
  xtick={0,.2,.4,.6,.8,1.0}, xticklabels={0,20,40,60,80,100},
  ytick={.68,.72,.76},
  yticklabel={\pgfmathprintnumber[fixed, fixed zerofill,
    precision=2]{\tick}},
  x tick label style={font=\scriptsize, gray!40!black},
  y tick label style={font=\scriptsize, gray!40!black},
  axis line style={gray!50},
  ymajorgrids, major grid style={gray!14},
  clip=false,
]
\draw[gray!45, densely dashed, line width=0.5pt]
  (axis cs:0.751,0.668) -- (axis cs:0.751,0.762);
\draw[gray!45, densely dashed, line width=0.5pt]
  (axis cs:-0.06,0.686) -- (axis cs:1.02,0.686);
\addplot[only marks, mark=triangle*, mark size=3.2pt,
  mark options={fill=gray!65, draw=white, line width=0.5pt}]
  coordinates {(0.751,0.686)};
\addplot[only marks, mark=diamond*, mark size=3.8pt,
  mark options={fill=figfullvta, draw=white, line width=0.5pt}]
  coordinates {(0.802,0.736)};
\addplot[only marks, mark=*, mark size=3.4pt,
  mark options={fill=figrepaironly, draw=white, line width=0.5pt}]
  coordinates {(0.030,0.723)};
\node[font=\scriptsize, anchor=east, fill=white, inner sep=1pt]
  at (axis cs:0.765,0.741) {full VTA (.736)};
\node[font=\scriptsize, anchor=south west, fill=white, inner sep=1pt]
  at (axis cs:0.055,0.722) {\textbf{repair-only} (.723)};
\node[font=\scriptsize, gray!25!black, anchor=south east,
  fill=white, inner sep=1pt]
  at (axis cs:0.71,0.6885) {predecessor (.686)};
\end{axis}
\end{tikzpicture}
\caption{Accuracy vs.\ conflict-free rewrite triggering ($y$-axis
zoomed; dashes cross at the predecessor, and up-left of them is better
on both axes). Repair-only improves on the predecessor on both ($3\%$
vs.\ $75\%$); full VTA adds the last $1.3$ points at $80\%$, mostly
\emph{verify} questions.}
\label{fig:pareto}
\end{figure}
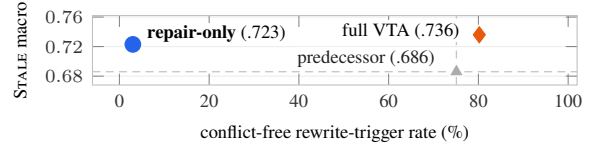

\paragraph{Overhead and latency, measured.}
Strict VTA costs $\approx$\$0.34/scenario (six calls, 9{,}328 audit
tokens; the scenario-joint variant \$\OffUsdPerScenario{}). Base and
transition audits and the three probes are independent, so the critical
path is three sequential calls --- a size-matched gateway probe measures
5.1/8.1\,s mean/P95 incremental latency (serving-stack, not a provider
guarantee).

\section{Limitations}\label{sec:limitations}
The $.879$ scenario-joint result is post-development and scope-confounded,
bundling evidence, reasoning, validation, and authority over all three
probes at once. The strict confirmation fixes the protocol; limits remain:
each arm is one sampled run ($n{=}100$ replicate $\approx$2-point shift);
proposer, adapter, and primary judge share the gpt-5.5 family, a
documented self-preference risk \citep{selfpref2026} the disjoint Gemini
reproduction only partially offsets; SR shows a T2
pocket (\StrictTTwoSrWins{}:\StrictTTwoSrLosses{}); and neither LongMemEval
nor natural traffic was rerun with strict. Human evaluation covers only
the scenario-joint ranking ($\kappa{=}\HumanKappa$). ``Verified'' means
provenance and chronology --- a provenance-valid path can still encode a
wrong state relation. A semantic check added as a second gate cuts the hard set's trap
over-repairs $\SemGateTrapVta{}{\to}\SemGateTrapSem{}$ while keeping true
ones, accuracy unchanged (supplement). That harder cross-family set
(\HardN{} scenarios) marks a boundary --- directives fire but give
\emph{no} accuracy gain (\HardMOneFull{}/\HardMOnePred{} deterministic,
\HardSemFull{}/\HardSemPred{} semantic, both n.s.), though over-correction
stays controlled; a powered human-authored set is still needed.
Finally, verbalizing sensitive transitions can entrench a wrong inference;
audit logs aid developers, not user correction, so deletion and correction
controls remain necessary.

\section{Conclusion}
Under \stale{}'s strict protocol, state-to-draft transition repair stably outperforms
a locked predecessor and a matched no-transition control
(\StrictBaseMacro{}$\to$\StrictMacro{}, IPA-led; disjoint judge
reproduced); a gold-store \horizonbench{} adaptation improves the
application of explicitly represented changes, while a harder authored
set yields no gain. This supports response repair in the studied
settings, not general-purpose memory updating.

\newpage
\bibliography{refs}


\end{document}